\documentclass[journal]{IEEEtran}
\usepackage{cite}

\ifCLASSINFOpdf
   \usepackage[pdftex]{graphicx}
\else
   \usepackage[dvips]{graphicx}
\fi
\usepackage{adjustbox}

\usepackage[cmex10]{amsmath}
\usepackage{bm}
\usepackage{amssymb}
\usepackage{algorithm,algorithmic}
\ifCLASSOPTIONcompsoc
  \usepackage[caption=false,font=normalsize,labelfont=sf,textfont=sf]{subfig}
\else
  \usepackage[caption=false,font=footnotesize]{subfig}
\fi
\usepackage{stfloats}
\usepackage{url}

\makeatletter
\newcommand{\defeq}{\mathrel{\aban@defeq}}
\newcommand{\aban@defeq}{%
  \vbox{\offinterlineskip\check@mathfonts
    \ialign{\hfil##\hfil\cr
      \fontsize{\ssf@size}{\z@}\normalfont def\cr
      \noalign{\kern1\p@}
      $\m@th=$\cr
      \noalign{\kern-.5\fontdimen22\textfont2}
    }%
  }%
}
\makeatother

\begin{document}

%
\title{Superpixel Segmentation Using Gaussian Mixture Model}
%
%
%
\author{Zhihua~Ban,
        Jianguo~Liu,~\IEEEmembership{Member,~IEEE,}
        and~Li~Cao%
\thanks{Z. Ban, J. Liu and L. Cao are with the 
National Key laboratory of Science and Technology on %
Multi-spectral Information Processing, School of Automation, %
Huazhong University of Science and Technology, Wuhan, Hubei Province 430074, %
China (e-mail: zhihua\_ban@hust.edu.cn; jgliu@ieee.org; caoli19871025@gmail.com).}}

\maketitle

\begin{abstract}
Superpixel segmentation algorithms are to partition an image into perceptually coherence atomic regions by assigning every pixel a superpixel label. Those algorithms have been wildly used as a preprocessing step in computer vision works, as they can enormously reduce the number of entries of subsequent algorithms. In this work, we propose an alternative superpixel segmentation method based on Gaussian mixture model (GMM) by assuming that each superpixel corresponds to a Gaussian distribution, and assuming that each pixel is generated by first randomly choosing one distribution from several Gaussian distributions which are defined to be related to that pixel, and then the pixel is drawn from the selected distribution. Based on this assumption, each pixel is supposed to be drawn from a mixture of Gaussian distributions with unknown parameters (GMM). An algorithm based on expectation-maximization method is applied to estimate the unknown parameters. Once the unknown parameters are obtained, the superpixel label of a pixel is determined by a posterior probability. The success of applying GMM to superpixel segmentation depends on the two major differences between the traditional GMM-based clustering and the proposed one: data points in our model may be non-identically distributed, and we present an approach to control the shape of the estimated Gaussian functions by adjusting their covariance matrices. Our method is of linear complexity with respect to the number of pixels. The proposed algorithm is inherently parallel and can get faster speed by adding simple OpenMP directives to our implementation. According to our experiments, our algorithm outperforms the state-of-the-art superpixel algorithms in accuracy and presents a competitive performance in computational efficiency.
\end{abstract}


\IEEEpeerreviewmaketitle

\section{Introduction}\label{sec:intro}

\IEEEPARstart{P}{artitioning} image into superpixels can be used as a preprocessing step for complex computer vision tasks, 
such as segmentation \cite{biseg,spbsegp,revrco1}, visual tracking \cite{tracking}, image matching \cite{Cheng20152269,Ma2015},
etc. Sophisticated algorithms benefit from working with superpixels, 
instead of just pixels, because superpixels reduce input entries and 
enable feature computation on more meaningful regions.

Like many terminologies in computer vision, there is no rigorous mathematical definition for superpixel. 
The commonly accepted description of a superpixel is ``a group of connected, perceptually homogeneous 
pixels which does not overlap any other superpixel.'' 
For superpixel segmentation, the following properties are generally desirable.

Prop. 1. \textbf{Accuracy}. Superpixels should adhere well to object boundaries. 
    Superpixels crossing object boundaries arbitrarily may lead to bad or catastrophic result for subsequent algorithms. \cite{VCells,seeds15,ERS,LSC}

Prop. 2. \textbf{Regularity}. The shape of superpixels should be regular. 
    Superpixels with regular shape make it easier to construct a graph for subsequent algorithms. 
    Moreover, these superpixels are visually pleasant which is helpful for algorithm designers' analysis. \cite{Polygons,LRW,lattices}

Prop. 3. \textbf{Similar size}. Superpixels should have a similar size. 
    This property enables subsequent algorithms to deal with each superpixel without bias 
    \cite{SLIC,Waterpixels,TP}. As pixels have the same ``size'' and the term of ``superpixel'' is originated from ``pixel'', 
    this property is also reasonable intuitively. This is a key property to distinguish between superpixel and other over-segmented regions.

Prop. 4. \textbf{Efficiency}. A superpixel algorithm should have a low complexity. 
    Extracting superpixels effectively is critical for real-time applications. \cite{SLIC,seeds15}.

\begin{figure*}[!htb]
\centering
\includegraphics[width=0.19\textwidth]{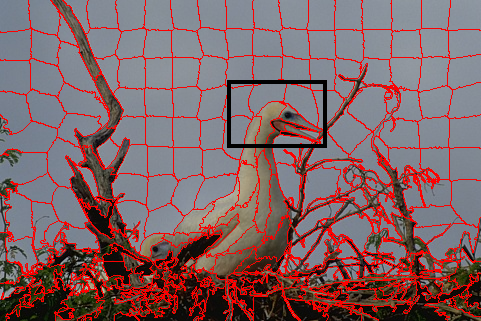}\ %
\includegraphics[width=0.19\textwidth]{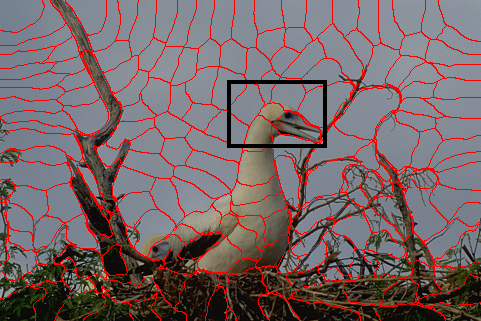}\ %
\includegraphics[width=0.19\textwidth]{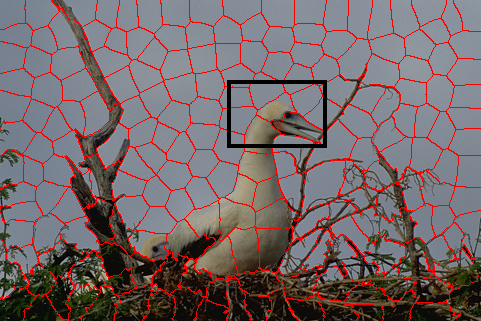}\ %
\includegraphics[width=0.19\textwidth]{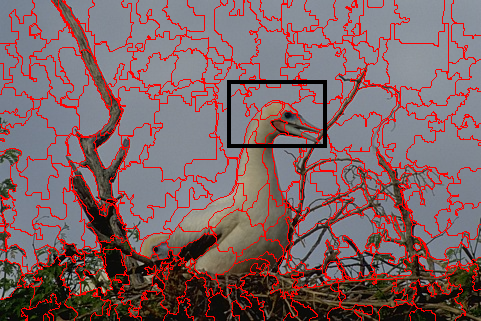}\ %
\includegraphics[width=0.19\textwidth]{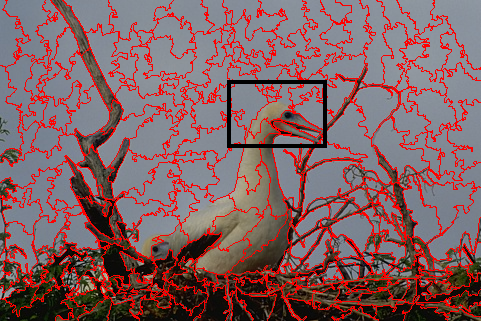}\\%
\vspace{1mm}
\includegraphics[width=0.19\textwidth]{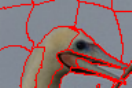}\ %
\includegraphics[width=0.19\textwidth]{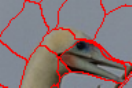}\ %
\includegraphics[width=0.19\textwidth]{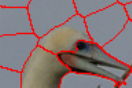}\ %
\includegraphics[width=0.19\textwidth]{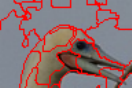}\ %
\includegraphics[width=0.19\textwidth]{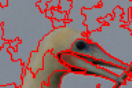}\\%
\vspace{-2.5mm}
\subfloat[]{\includegraphics[width=0.19\textwidth]{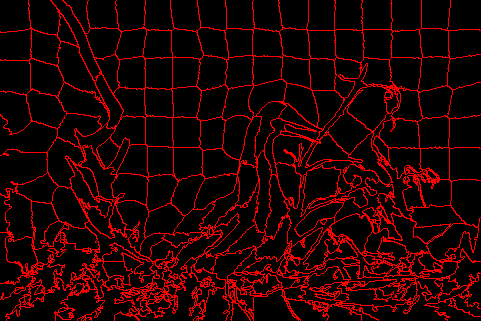}\label{fig:vc5:GMMSP}}\ %
\subfloat[]{\includegraphics[width=0.19\textwidth]{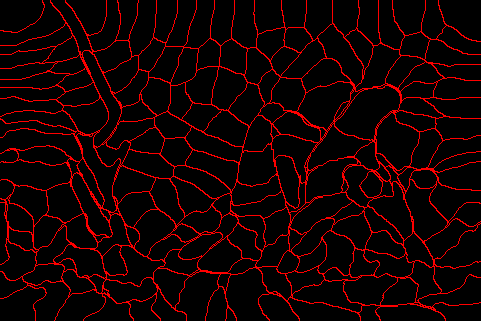}\label{fig:vc5:NC}}\ %
\subfloat[]{\includegraphics[width=0.19\textwidth]{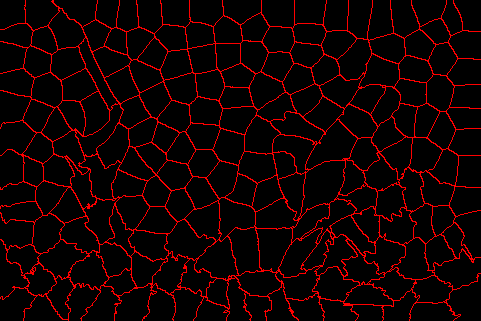}\label{fig:vc5:LRW}}\ %
\subfloat[]{\includegraphics[width=0.19\textwidth]{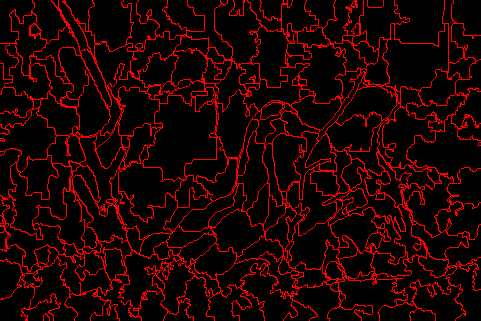}\label{fig:vc5:SEEDS}}\ %
\subfloat[]{\includegraphics[width=0.19\textwidth]{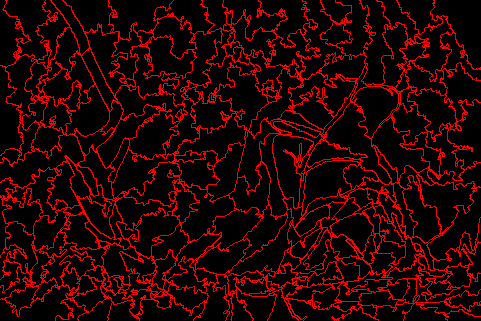}\label{fig:vc5:ERS}}\\%
\caption{Superpixel segmentations by five algorithms: \protect\subref{fig:vc5:GMMSP} Our method, \protect\subref{fig:vc5:NC} NC \cite{NC}, \protect\subref{fig:vc5:LRW} LRW \cite{LRW}, \protect\subref{fig:vc5:SEEDS} SEEDS \cite{seeds15}, and \protect\subref{fig:vc5:ERS} ERS \cite{ERS}. Each segmentation has approximately 200 superpixels. The second row zooms in the regions of interest defined by the white boxes in the first row. At the third row, superpixel boundaries are drawn to purely black images to highlight shapes of the superpixels.}
\label{fig:vc5}
\end{figure*}

Under the constraint of Prop. 3, the requirements on accuracy and regularity are to a certain extent oppositional. 
Intuitively, if a superpixel, with a limited size, needs to adhere well to object boundaries, 
the superpixel has to adjust its shape to that object which may be irregular. 
A satisfactory compromise between regularity and accuracy has not yet been found by existing superpixel algorithms.
As four typical algorithms shown in Fig. \ref{fig:vc5}\subref{fig:vc5:NC}-\ref{fig:vc5}\subref{fig:vc5:ERS}, 
the shape of superpixels generated by NC \cite{NC,NC_USE} (Fig. \ref{fig:vc5}\subref{fig:vc5:NC}) and LRW \cite{LRW} (Fig. \ref{fig:vc5}\subref{fig:vc5:LRW}) 
is more regular than that of superpixels extracted by SEEDS \cite{seeds15} (Fig. \ref{fig:vc5}\subref{fig:vc5:SEEDS}) 
and ERS \cite{ERS} (Fig. \ref{fig:vc5}\subref{fig:vc5:ERS}).
Nonetheless, the superpixels generated by SEEDS \cite{seeds15} and ERS \cite{ERS} adhere object boundaries better than those of NC \cite{NC} and LRW \cite{LRW}. 
In this work, A Gaussian mixture model (GMM) and an algorithm derived from the expectation-maximization algorithm \cite{EM} are built. 
It turns out the proposed method can strike a balance between regularity and accuracy. 
An example is displayed in Fig. \ref{fig:vc5}\subref{fig:vc5:GMMSP}, the compromise is that 
superpixels at regions with complex textures have an irregular shape to adhere object boundaries, while at homogeneous regions, the superpixels are regular.

Computational efficiency is a matter of both algorithmic complexity and implementation. 
Our algorithm has a linear complexity with respect to the number of pixels. 
As an algorithm has to read all pixels, linear time theoretically is the best time complexity for superpixel problem. 
Generally, algorithms can be categorized into two major groups: 
  parallel algorithms that are able to be implemented with parallel techniques and its performance scales with the number of parallel processing units, 
  and serial algorithms whose implementations are usually executed sequentially and only part of the system resources can be used on a parallel computer. 
Modern computer architectures are parallel and applications can benefit from parallel algorithms 
because parallel implementations generally run faster than serial implementations for the same algorithm. 
The proposed algorithm is inherently parallel and our serial implementation can easily achieve speedups by adding few simple OpenMP directives.

The proposed method is constructed by associating each superpixel to one Gaussian distribution;
modeling each pixel with a mixture of Gaussian distributions, which are related to the given pixel; 
and estimating unknown parameters in the proposed mixtures 
via an approach modified from the expectation-maximization algorithm;
The superpixel of a pixel is determined by a post probability.
The proposed approach was tested on the Berkeley Segmentation Data Set and Benchmarks 500 (BSDS500) \cite{BSDS500}. 
It is shown that the proposed method outperforms state-of-the-art methods in accuracy and presents a competitive performance in computational efficiency.
Our main contributions are summarized as follows:
\begin{enumerate}
  \item Our model is novel for superpixel segmentation, as GMM has not yet been well explored for the superpixel problem.
  \item We present a pixel-related GMM for each individual pixel, in which case pixels may be non-identically distributed, meaning that two pixels may have different GMMs.
  \item The proposed algorithm offers an option for controlling the regularity of superpixel shapes.
  \item Our algorithm is a parallel algorithm.
  \item The proposed approach give a better accuracy than state-of-the-art algorithms.
  \item Our method strike a balance between superpixel regularity and accuracy (see Fig. \ref{fig:vc5}\subref{fig:vc5:GMMSP}).
\end{enumerate}



The rest of this paper is organized as follows. 
Section \ref{sec:related} presents an overview of related works on superpixel segmentation. 
Section \ref{sec:method} introduces the proposed method. 
Experiments are discussed in section \ref{sec:exp}. Finally, the paper is concluded in section \ref{sec:cons}.

\section{Related works}\label{sec:related}
The concept of superpixel was first introduced by Xiaofeng Ren and Jitendra Malik in 2003 \cite{RenSeg}.
During the last decades, the superpixel problem has been well studied\cite{Peng15,Zhang16}. 
Existing superpixel algorithms extract superpixels either by optimizing superpixel boundaries, 
such as finding paths and evolving curves, or by grouping pixels, e.g. the most well-known SLIC \cite{SLIC}.
%
%
We will give a brief review on how existing algorithms solve the superpixel problem in the two aspects in this section.

\emph{Optimize boundaries}.
Algorithms extract superpixels not by labeling pixels directly but by marking superpixel boundaries, 
or by only updating the label of pixels on superpixel boundary is in this category.
Rohkohl et al. present a superpixel method that iteratively assigns superpixel boundaries to their most
similar neighboring superpixel \cite{rohkohl2007efficient}. 
A superpixel is represented with a group of pixels that are randomly selected from that superpixel. 
The similarity between a pixel and a super-pixel is defined as the average similarities 
from the pixel to all the selected representatives.
Aiming to extract lattice-like superpixels, or ``superpixel lattices'', 
\cite{lattices} partitions an image into superpixels by gradually adding horizontal and vertical 
paths in strips of a pre-computed boundary map. 
The paths are formed by two different methods: s-t min-cut and dynamic programming. 
The former finds paths by graph cuts and the latter constructs paths directly. 
The paths have been designed to avoid parallel paths crossing and guarantee 
perpendicular paths cross only once. 
The idea of modeling superpixel boundaries as paths (or seam carving \cite{seamcarv}) 
and the use of dynamic programming were borrowed by later variations or 
improvements \cite{zhu2015fast,latticecut,RPSS,topology,gridseams,returnseams}.
In TurboPixels \cite{TP}, Levinshtein et al. model the boundary of each superpixel as a closed curve. 
So, the connectivity is naturally guaranteed. Based on level-set evolution, the curves gradually sweep
over the unlabeled pixels to form superpixels under the constraints of two velocities. 
%
In VCells \cite{VCells}, a superpixel is represented as a mean vector 
of color of pixels in that superpixel. With the designed distance \cite{VCells}, 
VCells iteratively updates superpixel boundaries to their nearest neighboring superpixel. 
The iteration stops when there are no more pixels need to be updated.
SEEDS \cite{seeds12,seeds15} exchanges superpixel boundaries using a hierarchical structure. 
At the first iteration, the biggest blocks on superpixel boundary are updated for a better energy. 
The size of pixel blocks becomes smaller and smaller as the number of iterations increases. 
The iteration stops after the update of boundary exchanges in pixel level.
Improved from SLIC \cite{SLIC}, \cite{yamaguchi2014efficient} and \cite{yao2015real} 
present more complex energy. To minimize their corresponding energy, 
\cite{yamaguchi2014efficient} and \cite{yao2015real} update boundary pixels instead of assigning 
a label for all pixels in each iteration. Based on \cite{yamaguchi2014efficient}, 
\cite{yao2015real} adds the connectivity and superpixel size into their energy. 
For the pixel updating, \cite{yao2015real} uses a hierarchical structure 
like SEEDS \cite{seeds12}, while \cite{yao2015real} exchanges labels only in pixel level. 
Zhu et al. propose a speedup of SLIC \cite{SLIC} by only moving unstable boundary pixels, 
the label of which changed in the previous iteration \cite{zhu2015fast}.
Besides, based on pre-computed line segments or edge maps of the input image, 
\cite{li2015edge} and \cite{Polygons} extract superpixels by aligning superpixel boundaries to the lines or the edges.

\emph{Grouping pixels}.
Superpixels algorithms that assign labels for all pixels in each iteration is in this category.
With an affinity matrix constructed based on boundary cue \cite{Martin04}, 
the algorithm developed in \cite{NC_USE}\cite{RenSeg}, which is usually abbreviated as NC \cite{SLIC}, 
uses normalized cut \cite{NC} to extract superpixels. 
%
In Quick shift (QS) \cite{QS}, the pixel density is estimated on a Parzen window with a Gaussian kernel. 
A pixel is assigned to the same group with its parent which is the nearest pixel with a greater density 
and within a specified distance. QS does not guarantee connectivity, or in other words, pixels with the same label may not be connected.
Veksler et al. propose an approach that distributes a number of overlapping square patches 
on the input image and extracts superpixels by finding a label for each pixel from patches 
that cover the present pixel \cite{GC}. The expansion algorithm in \cite{boykov2001fast} is 
gradually adapted to modify pixel label within local regions with a fixed size in each iteration. 
A similar solution in \cite{zhang2011superpixels} is to formulate the superpixel problem 
as a two-label problem and build an algorithm through grouping pixels into vertical and horizontal 
bands. By doing this, pixels in the same vertical and horizontal group form a superpixel.
Starting from an empty graph edge set, ERS \cite{ERS} sequentially adds edges to the set until the 
desired number of superpixels is reached. At each adding, ERS \cite{ERS} takes the edge that 
results in the greatest increase of an objective function. The number of generated superpixels 
is exactly equal to the desired number. 
SLIC \cite{SLIC} is the most well-known superpixel algorithm due to its efficiency and simplicity. 
In SLIC \cite{SLIC}, a pixel corresponds to a five dimensional vector including color and spatial location, 
and $k$-means is employed to cluster those vectors locally, i.e. each pixel only compares 
with superpixels that fall into a specified spatial distance and is assigned to the nearest superpixel. 
Many variations follow the idea of SLIC in order to either decrease its run-time \cite{compact,onepass,gslic} 
or improve its accuracy \cite{nslic,yamaguchi2014efficient}. 
LSC \cite{LSC} also uses a $k$-means method to refine superpixels. Instead of directly using the 
5D vector used in SLIC \cite{SLIC}, LSC \cite{LSC,Ban2016} maps them to a feature space and a weighted 
$k$-means is adopted to extract superpixels. 
Based on marker-based watershed transform, \cite{Waterpixels} and \cite{compact} incorporate spatial 
constraints to an image gradient in order to produce superpixels with regular shape and similar size. 
%
LRW \cite{LRW} groups pixels using an improved random walk algorithm. 
By using texture features to optimize an initial superpixel map, this method can produce regular superpixels in regions with complex texture. However, this method suffers from a very slow speed.

Although FH \cite{FH}, mean shift \cite{MS} and watersheds \cite{Watersheds1991}, 
have been refereed to as ``superpixel'' algorithms in the literature, they are not 
covered in this paper as the sizes of the regions produced by them vary enormously. 
This is mainly because these algorithms do not offer direct control to the size of the 
segmented regions. Structure-sensitive or content-sensitive superpixels in 
\cite{MSLIC,SSS} are also not considered to be superpixels, 
as they do not aim to extract regions with similar size (see Prop. 3 in section \ref{sec:intro}).

A large number of superpixel algorithms have been proposed, however, few models have been presented
and most of the existing energy functions are variation of the objective function of 
$k$-means. In our work, we propose an alternative model to tackle the superpixel problem. 
With an elaborately designed algorithm, the underlying segmentation from the model is well revealed.


\section{The method}\label{sec:method}

\subsection{Model}

Let $i$ stands for the pixel index of an input image 
$I$ with its width $W$ and height $H$ in pixels. 
Hence, the total number of pixels $N$ of image $I$ is $W\cdot H$, and $i \in V \defeq \{0,1,\ldots,N-1\}$.
Let $(x_i, y_i)$ denotes pixel $i$'s position on the image plane, where $x_i \in \{0,1,\ldots,W-1\}$
and $y_i \in \{0,1,\dots,H-1\}$, and $c_i$ denotes pixel $i$'s intensity or color. 
If color image is used, $c_i$ is a vector, otherwise, $c_i$ is a scalar. 
The number of elements in $c_i$ is ignored for now and it will be discussed later.
We use vector $\mathbf{z}_i = (x_i,y_i,c_i)^T$ to represent pixel $i$.


Most existing superpixel algorithms require the desired number of superpixels $K$ as an input.
However, instead of using $K$ directly, we use $v_x$ and $v_y$ as essential inputs.
If $K$ is specified, $v_x$ and $v_y$ are obtained by the following equation.
\begin{equation}\label{eq:vxy}
v_x = v_y = \bigg\lfloor \sqrt{\frac{W\cdot H}{K}} \bigg\rfloor\,.
\end{equation}
If $v_x$ and $v_y$ are preferred, it is encouraged to assign the same value to the two variables.
Using equation \eqref{eq:nxyK}, the desired number of superpixels $K$ is computed when $v_x$ and $v_y$ are directly specified,
or re-computed in the case when $v_x$ and $v_y$ are obtained by equation \eqref{eq:vxy}.
\begin{equation} \label{eq:nxyK}
n_x = \bigg\lfloor \frac{W}{v_x} \bigg\rfloor ,\, \ n_y = \bigg\lfloor \frac{H}{v_y} \bigg\rfloor ,\, K = n_x \cdot n_y \,.
\end{equation}
For simplicity of discussion, we assume that $W\bmod v_x=0$ and $H \bmod v_y = 0$. 
We define the superpixel set as $\mathcal{K} \defeq \{0,1,\ldots,K-1\}$.

Each superpixel $k\in \mathcal{K}$ corresponds to a Gaussian distribution with p.d.f. $p(\mathbf{z}; \bm\theta_k)$, where
$\bm\theta_k = \{\bm\mu_k, \bm\Sigma_k\}$ and 
\begin{align}
&p(\mathbf{z}; \bm\theta_k) = \label{eq:gaussian}\\
&\frac{1}{(2\pi)^{D/2} \sqrt{\det(\bm\Sigma_k)}} \exp\bigg\{-\frac{1}{2} (\mathbf{z}-\bm{\mu}_k)^T \bm\Sigma_k^{-1} (\mathbf{z}-\bm{\mu}_k) \bigg\}\,,\nonumber
\end{align}
in which $D$ is the number of components in $\mathbf{z}$.

\begin{figure}[!b]
\centering
\includegraphics{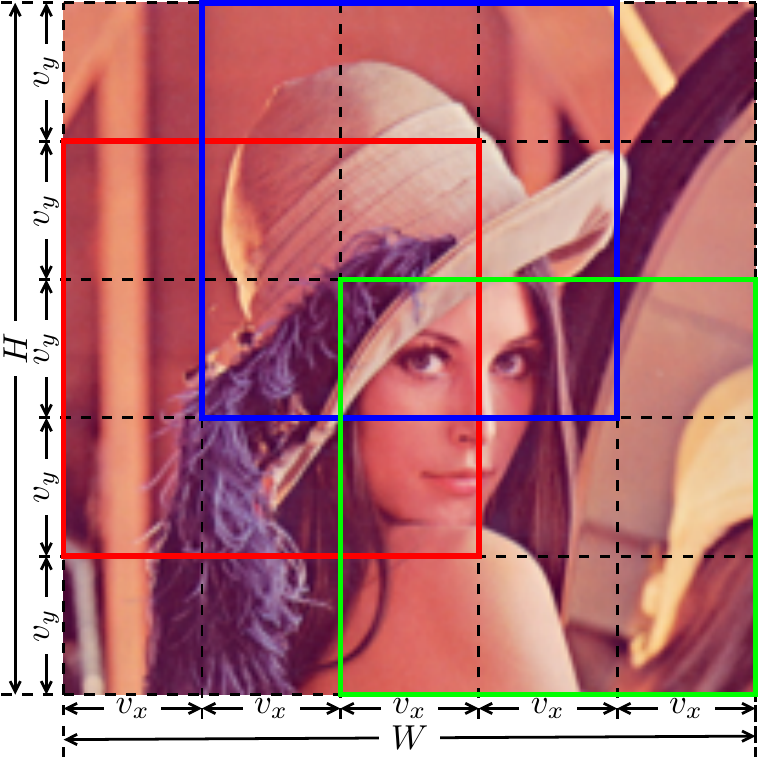}
\caption{Illustration of pixel set $I_k$. 
Pixel set $I_{7}$, $I_{11}$, and $I_{18}$ are correspondingly surrounded 
with blue, red, and green rectangles in this figure.
}\label{fig:illusIk}
\end{figure}

If pixel $i$ is drawn from superpixel $k$, we assume that pixel $i$ can be only in pixel set $I_k$ which is defined in equation \eqref{eq:Ik}.
Fig. \ref{fig:illusIk} gives an visual illustration for $I_k$. The definition of $I_k$ is one of the key points in our method.
\begin{equation}
I_k \defeq \{i\ |\ x_{k,b} \leq x_i <  x_{k,e}, y_{k,b} \leq y_i <  y_{k,e}, i \in {I} \}\,, \label{eq:Ik}
\end{equation}
where
\begin{IEEEeqnarray}{rCl}
x_{k,b} &\defeq& \max(0, v_x\cdot(k_x-1)) \,, \\
x_{k,e} &\defeq& \min(W, v_x\cdot(k_x+2)) \,, \\
y_{k,b} &\defeq& \max(0, v_y\cdot(k_y-1)) \,, \\
y_{k,e} &\defeq& \min(H, v_y\cdot(k_y+2)) \,,
\end{IEEEeqnarray}
and for any given superpixel $k \in \mathcal{K}$, we have
\begin{equation}\label{eq:kxy}
k_x \defeq k \bmod n_x\,,\, k_y \defeq \lfloor k / n_x \rfloor.
\end{equation}
For each pixel $i\in V$, the possible superpixels from which pixel $i$ may be generated 
form a superpixel set $K_i \defeq \{k | i \in I_k, k \in \mathcal{K}\} \subseteq \mathcal{K}$.
Let $\tilde{L}_i$ stand for the unknown superpixel label of pixel $i$, and 
$\tilde{L}_i$ are treated as random variables whose possible values are in $K_i$, $i \in V$.
We now treat $\mathbf{z}_i$ as observations of random variables $\mathbf{Z}_i$.
The probability density function $p_i(\mathbf{z})$ of each random variables $\mathbf{Z}_i$ is defined as 
a mixture of Gaussian functions, known as Gaussian mixture model (GMM). 
\begin{equation}\label{eq:pzi}
p_i(\mathbf{z}) = \sum_{k \in K_i} \Pr (\tilde{L}_i=k) p(\mathbf{z}; \bm\theta_k)\,, \forall i \in V\,,
\end{equation}
in which $\Pr (\tilde{L}_i=k)$, the probability that $\tilde{L}_i$ takes value $k$, 
are defined to be $P_i \defeq 1/|K_i|$ for $k \in K_i$, where $|\cdot|$ is the number of elements in a given set.
Therefore, $p_i(\mathbf{z})$ become
\begin{equation}
p_i(\mathbf{z}) = P_i \sum_{k \in K_i} p(\mathbf{z}; \bm\theta_k)\,.
\end{equation}
Note that pixels may have different distributions when $K_i\subsetneqq \mathcal{K}$ which is the most common case.
This is the main difference between our GMM and the traditional GMM. The usage of $P_i$ results in superpixels with similar size.

Once an estimator of $\bm\theta \defeq \{\bm\theta_k | k \in \mathcal{K}\}$ is found, superpixel label $L_i$ of pixel $i$
can be obtained by 
\begin{equation}
L_i = \arg_k \max_{k\in K_i} \Pr(\tilde{L}_i=k | \mathbf{Z}_i = \mathbf{z}_i)\,,
\end{equation}
By Bayes' theorem, we have the posterior probability of each $\tilde{L}_i$,
\begin{IEEEeqnarray}{l}
\Pr(\tilde{L}_i=k | \mathbf{Z}_i = \mathbf{z}_i) = \IEEEnonumber\\
\frac{p(\mathbf{z}_i;\bm{\theta}_k)\Pr(\tilde{L}_i = k)}{\sum_{k \in K_i}p(\mathbf{z}_i;\bm{\theta}_k)\Pr(\tilde{L}_i = k)} 
= \frac{p(\mathbf{z}_i;\bm{\theta}_k)}{\sum_{k \in K_i}p(\mathbf{z}_i;\bm{\theta}_k)}\,. \label{eq:post}
\end{IEEEeqnarray}
Therefore, superpixel labels can be obtained by 
\begin{equation}\label{eq:obtLi}
L_i = \arg_k \max_{k\in K_i} \frac{p(\mathbf{z}_i;\bm{\theta}_k)}{\sum_{k \in K_i}p(\mathbf{z}_i;\bm{\theta}_k)}\,.
\end{equation}

\subsection{Parameter estimation} \label{sec:estimate}

Maximum likelihood estimation is used to estimate the parameters in $\bm\theta$. 
Suppose that $\mathbf{Z}_i$, $i\in V$, are independently distributed.
For all observed vectors $\mathbf{z}_i$, $i\in V$, the logarithmic likelihood function will be 
\begin{IEEEeqnarray}{rCl}
f(\bm\theta) &=& \sum_{i\in V} \ln p_i(\mathbf{z}_i)\IEEEnonumber\\
             &=& \sum_{i\in V} \ln P_i + \sum_{i\in V} \ln \sum_{k\in K_i} p(\mathbf{z}_i; \bm\theta_k)\,. \label{eq:ftheta}
\end{IEEEeqnarray}
Because $\sum_{i\in V} \log P_i$ is constant, the value of $\bm\theta$ that maximizes $f(\bm\theta)$ will be the same as the value of $\bm\theta$
that maximizes 
\begin{equation}
L(\bm\theta) = \sum_{i\in V} \ln \sum_{k\in K_i} p(\mathbf{z}_i; \bm\theta_k)\,.
\end{equation}

According to Jensen's inequality, $L(\bm\theta)$ is greater than or equal to $Q(\bm{R},\bm{\theta})$ as shown below.
\begin{IEEEeqnarray}{rCl}
L(\bm{\theta}) &=& \sum_{i\in V}\ln \sum_{k \in K_i} R_{i,k} \frac{p(\mathbf{z}_i; \bm{\theta}_k)}{R_{i,k}} \label{eq:LR}\\
&\geq& \sum_{i\in V} \sum_{k \in K_i} R_{i,k} \ln \frac{p(\mathbf{z}_i; \bm{\theta}_k)}{R_{i,k}} = Q(\bm{R},\bm{\theta})\,,\label{eq:Q}
\end{IEEEeqnarray}
where $R_{i,k} \geq 0$, $\sum_{k\in K_i} R_{i,k} = 1$ for $i \in V$ and $k \in K_i$, and $\bm{R}=\{R_{i,k}\ |\ i\in V, k\in K_i\}$.
We now use the expectation-maximization algorithm to iteratively find the value of $\bm\theta$ that maximizes $Q(\bm{R},\bm{\theta})$ 
to approach the maximum of $L(\bm{\theta})$ with two steps: the expectation step (E-step) and the maximization step (M-step).

\emph{E-step}: once a guess of $\bm{\theta}$ is given, $Q(\bm{R},\bm{\theta})$ is 
expected to be tightly attached to $L(\bm{\theta})$. 
To this end, $\bm{R}$ is required to ensure $L(\bm{\theta}) = Q(\bm{R},\bm{\theta})$. 
Equation \eqref{eq:Rconfig} is a sufficient condition for Jensen's inequality to hold the equality of inequality $L(\bm\theta)\geq Q(\bm{R},\bm{\theta})$.
\begin{equation}\label{eq:Rconfig}
\frac{p(\mathbf{z}_i; \bm{\theta}_k)}{R_{i,k}} = \alpha\,,
\end{equation}
where $\alpha$ is a constant. Since $\sum_{k\in K_i} R_{i,k} = 1$, 
$\alpha$ can be eliminated and hence $R_{i,k}$ can be updated by equation \eqref{eq:Rik}
to hold the equality to be true.
\begin{equation}\label{eq:Rik}
R_{i,k} = \frac{p(\mathbf{z}_i;\bm{\theta}_k)}{\sum_{k \in K_i}p(\mathbf{z}_i;\bm{\theta}_k)} \,.
\end{equation}

\emph{M-step}: in this step, 
$\bm{\theta}$ is derived by maximizing $Q(\bm{R},\bm{\theta})$ with a given $\bm{R}$. 
To do this, we first calculate the derivatives of $Q(\bm{R},\bm{\theta})$ with respect to mean vectors $\bm{\mu}_k$ and covariance matrices $\bm\Sigma_k$, 
and set the derivatives to zero, as shown in equations \eqref{eq:derMu}-\eqref{eq:derSys}. 
Then the parameters are obtained by solving equation \eqref{eq:derSys}. 
\begin{equation}\label{eq:derMu}
\frac{\partial Q(\bm{R},\bm{\theta})}{\partial \bm{\mu}_k} = 
\sum_{i\in I_k} R_{i,k} \bigg\{ \bm\Sigma_k^{-1} (\mathbf{z}_i - \bm{\mu}_k)\bigg\}\text{,}
\end{equation}
\begin{IEEEeqnarray}{l}\label{eq:derSig}
\hspace{-1em}\frac{\partial Q(\bm{R},\bm{\theta})}{\partial \bm\Sigma_k} = \IEEEnonumber\\
\hspace{-1em}\sum_{i\in I_k} R_{i,k} \bigg\{ \frac{1}{2}\bm\Sigma_k^{-1} (\mathbf{z}_i - \bm{\mu}_k)(\mathbf{z}_i - \bm{\mu}_k)^{T} \bm\Sigma_k^{-1} - \frac{1}{2}\bm\Sigma_k^{-1} \bigg\} ,
\end{IEEEeqnarray}
\begin{equation}\label{eq:derSys}
\frac{\partial Q(\bm{R},\bm{\theta})}{\partial \bm{\mu}_k} = \bm0 ,\ 
\frac{\partial Q(\bm{R},\bm{\theta})}{\partial \bm\Sigma_k} = \bm0,
\end{equation}
\begin{equation}\label{eq:UpdateMu}
\bm{\mu}_k = \frac{\sum_{i\in I_k} R_{i,k} \mathbf{z}_i}{\sum_{i\in I_k} R_{i,k}}\ \text{,}
\end{equation}
\begin{equation}\label{eq:UpdateSig}
\bm\Sigma_k = \frac{\sum_{i\in I_k} R_{i,k}(\mathbf{z}_i-\bm{\mu}_k)(\mathbf{z}_i-\bm{\mu}_k)^T}{\sum_{i\in I_k} R_{i,k}} \,.
\end{equation}

After initializing $\bm\theta$, the estimate of $\bm\theta$ is 
obtained by iteratively updating $\bm{R}$ and $\bm\theta$ using equations 
\eqref{eq:Rik}, \eqref{eq:UpdateMu}, and \eqref{eq:UpdateSig} 
until $\bm\theta$ converges.

\subsection{Algorithm in practice}\label{sec:algoprac}

Although the estimate of $\bm\theta$ in section \ref{sec:estimate} 
supports full covariance matrices, i.e., a covariance matrix with all its 
elements as shown in equation \eqref{eq:UpdateSig}, only block diagonal 
matrices are used in this work (see equation \eqref{eq:BlockSigma}). 
This is because computing on block diagonal matrices 
is more efficient than computing on full matrices, and
full matrices will also not bring better performance in accuracy.
\begin{equation}\label{eq:BlockSigma}
\bm\Sigma_{k} = 
\begin{bmatrix}
  \bm\Sigma_{k,s} & \bm{0} \\
  \bm{0}          & \bm\Sigma_{k,c}
\end{bmatrix}\text{,}
\end{equation}
where $\bm\Sigma_{k,s}$ and $\bm\Sigma_{k,c}$ respectively represent the 
spatial covariance matrices and the color covariance matrices for $k \in \mathcal{K}$. 
For color images, it is encouraged to split their color covariance matrices into 
lower dimensional matrices to save computation. 
For example, if an image with CIELAB color space is inputted, 
it is better to put color-opponent dimensions $a$ and $b$ into a 2 by 2 covariance matrix. 
In this case, $\bm\Sigma_{k,c}$ in equation \eqref{eq:BlockSigma} will become
\begin{equation}\label{eq:BlockSigma1}
\bm\Sigma_{k,c} = 
\begin{bmatrix}
  \sigma_{k,l}^2 & \bm{0}      \\
  \bm{0}         & \bm\Sigma_{k,(a,b)}
\end{bmatrix}.
\end{equation}
However, we will keep using \eqref{eq:BlockSigma} to discuss the proposed algorithm for simplicity.

The covariance matrices will be updated according to equations 
\eqref{eq:UpdateSigS} and \eqref{eq:UpdateSigC} which are derived 
by replacing $\bm\Sigma_k$ in equation \eqref{eq:derSig} with the block 
diagonal matrices in equation \eqref{eq:BlockSigma}, and by further solving \eqref{eq:derSys}.
\begin{equation}\label{eq:UpdateSigS}
\bm\Sigma_{k,s} = \frac{\sum_{i\in I_k} R_{i,k}(\mathbf{z}_{i,s}-\bm{\mu}_{k,s})(\mathbf{z}_{i,s}-\bm{\mu}_{k,s})^T}{\sum_{i\in I_k} R_{i,k}} \text{,}
\end{equation}
\begin{equation}\label{eq:UpdateSigC}
\bm\Sigma_{k,c} = \frac{\sum_{i\in I_k} R_{i,k}(\mathbf{z}_{i,c}-\bm{\mu}_{k,c})(\mathbf{z}_{i,c}-\bm{\mu}_{k,c})^T}{\sum_{i\in I_k} R_{i,k}} \text{,}
\end{equation}
where $\mathbf{z}_{i,s}$ and $\bm{\mu}_{i,s}$ are the spatial components 
of $\mathbf{z}_i$ and $\bm{\mu}_i$, and $\mathbf{z}_{i,c}$ and $\bm{\mu}_{i,c}$ are, 
for grayscale images, the intensity components, or, for color image, the color components of $\mathbf{z}_i$ and $\bm{\mu}_i$. 

Since $\bm\Sigma_{k,s}$ and $\bm\Sigma_{k,c}$ are positive semi-definite in practice, 
they may be not invertible sometimes. To avoid this trouble, we first compute the 
eigendecompositions of the covariance matrices as shown in equations \eqref{eq:eigdecs} 
and \eqref{eq:eigdecc}, then eigenvalues on the major diagonals of $\Lambda_{k,s}$ and 
$\Lambda_{k,c}$ are modified using equations \eqref{eq:mlams} and  \eqref{eq:mlamc}, and 
finally $\bm\Sigma_{k,s}$ and $\bm\Sigma_{k,c}$ are reconstructed via the equations \eqref{eq:eigdecrs} and \eqref{eq:eigdecrc}.
\begin{IEEEeqnarray}{rCl}
\bm\Sigma_{k,s} &=& Q_{k,s}\ \Lambda_{k,s} \ Q_{k,s}^{-1} \,,\label{eq:eigdecs}\\
\bm\Sigma_{k,c} &=& Q_{k,c}\ \Lambda_{k,c} \ Q_{k,c}^{-1} \,,\label{eq:eigdecc}
\end{IEEEeqnarray}
where $\Lambda_{k,s}$ and $\Lambda_{k,c}$ are diagonal matrices with eigenvalues on their respective major diagonals, 
and $Q_{k,s}$ and $Q_{k,c}$ are orthogonal matrices. 
We use $\lambda_{k,s}(j_s)$ and $\lambda_{k,c}(j_c)$ to denote the respective eigenvalues on major diagonals of $\Lambda_{k,s}$ and $\Lambda_{k,c}$, 
where $j_s\in\{0,1\}$ and $j_c\in \{0,1,2\}$.
If the input image is grayscale, then we will have that $Q_{k,c}=1$, $\bm\Sigma_{k,c}$ and $\Lambda_{k,c}$ are scalars, and $j_c = 0$.
\begin{equation}\label{eq:mlams}
\tilde{\lambda}_{k,s}(j_s) = \left\{
\begin{array}{rl}
\lambda_{k,s} & \text{if } \lambda_{k,s}(j_s) \geq \epsilon_s \,,\\
\epsilon_s          & \text{else}.
\end{array}
\right.
\end{equation}
\begin{equation}\label{eq:mlamc}
\tilde{\lambda}_{k,c}(j_c) = \left\{
\begin{array}{rl}
\lambda_{k,c} & \text{if } \lambda_{k,c}(j_c) \geq \epsilon_c \,,\\
\epsilon_c          & \text{else}. 
\end{array}
\right.
\end{equation}
where $\epsilon_s$ and $\epsilon_c$ are two constants.
Although this two constants are originally designed to prevent covariance matrices from being singular, they also give
an opportunity to control regularity of the generated superpixels by weighing the relative importance between spatial proximity and color similarity.
For instance, a larger $\epsilon_c$ produces more regular superpixels, and the opposite is true for a smaller $\epsilon_c$. 
As $\epsilon_c$ and $\epsilon_s$ are opposite to each other, we set $\epsilon_s=2$ and leave  $\epsilon_c$ for detailed description in section \ref{sec:exp}.
\begin{IEEEeqnarray}{rCl}
\bm\Sigma_{k,s} = Q_{k,s}\ \tilde{\Lambda}_{k,s} \ Q_{k,s}^{-1} \,,\label{eq:eigdecrs}\\
\bm\Sigma_{k,c} = Q_{k,c}\ \tilde{\Lambda}_{k,c} \ Q_{k,c}^{-1} \,,\label{eq:eigdecrc}
\end{IEEEeqnarray}
where $\tilde{\Lambda}_{k,s}$ and $\tilde{\Lambda}_{k,c}$ are diagonal matrices 
with $\tilde{\lambda}_{k,s}(j_s)$ and $\tilde{\lambda}_{k,c}(j_c)$ on their respective major diagonals.

In the proposed algorithm, $\bm\mu_k$ are initialized using $K$ center pixels over the 
input image uniformly at fixed horizontal and vertical intervals $v_x$ and $v_y$,
i.e. $\bm\mu_k = \mathbf{z}_j$, where 
\begin{equation}
j = k_x \cdot v_x + \lfloor v_x/2\rfloor + W\cdot(k_y \cdot v_y + \lfloor v_y/2\rfloor).
\end{equation}
We initialize $\bm\Sigma_{k,s}$ with $\text{diag}(v_x^2, v_y^2)$ so that neighboring superpixels can be well overlapped at the beginning.
The initialization of $\bm\Sigma_{k,c}$ is not very straightforward, the basic idea is to set their main diagonal equal to the square of a small color distance $\lambda$
with which two pixels are perceptually uniform. The effect of different values for $\lambda$ will be discussed in section \ref{sec:exp}.

Once parameter $\bm\theta$ is initialized, it will finally be estimated by iteratively updating 
\eqref{eq:Rik}, \eqref{eq:UpdateMu}, \eqref{eq:eigdecrs}, and \eqref{eq:eigdecrc} until $\bm\theta$ converges.
As a preprocessing step to subsequent applications, superpixel algorithm should run as fast as possible.
We have found that iterating 10 times is sufficient for most images without checking convergence,
and we will use this iteration number for all our experiments and will denote it with $T$ to avoid confusion.

As the connectivity of superpixels cannot be guaranteed, a postprocessing step is required to enforce connectivity of the generated superpixels. 
This is done by sorting the isolated superpixels in ascending order according to their sizes, 
and sequentially merging small isolated superpixels, which are less than one fourth of the 
desired superpixel size, to their nearest neighboring superpixels, with only intensity or color being taken into account.
Once an isolated superpixel (source) is merged to another superpixel (destination), 
the size of the source superpixel is cleared to zero, and the size of the destination superpixel 
will be updated by adding the size of the source superpixel. 
This size updating trick will prevent the size of the produced superpixels from significantly varying.

The proposed algorithm is summarized in Algorithm 1.
\begin{algorithm}[H]
\caption{The proposed superpixel algorithm.}
\begin{algorithmic}[1]
\REQUIRE $v_x$ and $v_y$, or $K$; $I$.
\ENSURE  $L_i$, $i\in V$.
\STATE Initialize parameter $\bm\theta$.
\STATE Update $\bm{R}$ using equation \eqref{eq:Rik}, and set $t=0$.
\WHILE{$t<T$}
 \STATE Update $\bm{\mu}_k$ using equation \eqref{eq:UpdateMu}.
 \STATE Update $\bm\Sigma_{k,s}$ and $\bm\Sigma_{k,c}$ using equations \eqref{eq:eigdecrs} and \eqref{eq:eigdecrc}.
 \STATE Update $\bm{R}$ using equation \eqref{eq:Rik}, and set $t=t+1$.
\ENDWHILE
\STATE $L_i$ are determined by equation \eqref{eq:obtLi}. 
\STATE Postprocessing for connectivity enforcement.
\end{algorithmic}
\end{algorithm}

\subsection{Analysis on the proposed method}\label{sec:analysis}

As the frequency of a single processor is difficult to improve, 
modern processors are designed using parallel architectures. 
If an algorithm is able to be implemented with parallel techniques,
its performance generally scales with the number of parallel processing units and its computational 
efficiency can be significantly improved on multi-core or on many-core systems. 
Fortunately, the most expensive part of our algorithm, namely the iteration of updating of $\bm{R}$ and $\bm{\theta}$, can be parallelly executed
as each $R_{i,k}$ can be updated independently, and so do $\bm{\mu}_k$ and $\bm\Sigma_k$. 
In our experiments, we will show that our C++ implementation is easy to get speedup on multi-core CPUs with only few OpenMP directives inserted.

By the definition of $K_i$, we have $1 \leq |K_i| \leq 9$ for $i \in V$.
Therefore, the updating of $\bm{R}$ has a complexity of $\mathcal{O}((T+1) \cdot N)$.
Because we use $T$ as a constant in the proposed algorithm, the complexity of $\bm{R}$ is $\mathcal{O}(N)$.
By the definition of $I_k$, we have $v_x\cdot v_y \leq |I_k| \leq 9 \cdot v_x \cdot v_y$.
Based on equations \eqref{eq:UpdateMu}, \eqref{eq:UpdateSigS}, and \eqref{eq:UpdateSigC}, 
the complexity of updating $\bm{\theta}$ is $\mathcal{O}(T \cdot K\cdot |I_k|)$. 
Since $K\cdot |I_k| = (n_x \cdot n_y) \cdot (v_x \cdot v_y) = W\cdot H = N$,
the updating of $\bm\theta$ has a complexity of $\mathcal{O}(N)$.
In the worst case, the sorting procedure in the postprocessing step requires $\mathcal{O}(m^2)$ operations, 
where $m$ is the number of isolated superpixels. The merging step needs $\mathcal{O}(\tilde{m}\cdot n)$ operations, 
where $\tilde{m}$ is the number of small isolated superpixels and $n$ represents the average number of their adjacent neighbors. 
In practice, $m^2 +\tilde{m}\cdot n \ll T\cdot N$, the operations required for the postprocessing step can be ignored. 
Therefore, the proposed superpixel algorithm is of a linear complexity $\mathcal{O}(N)$.

\begin{figure*}[!tb]
\centering
\subfloat[]{\includegraphics[width=0.31\linewidth]{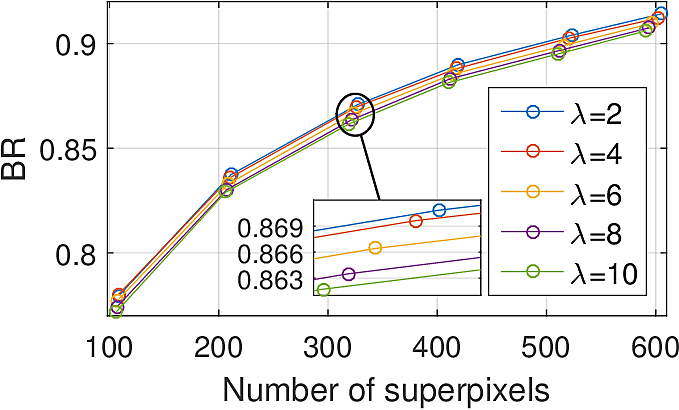}\label{fig:lbr}} \quad
\subfloat[]{\includegraphics[width=0.31\linewidth]{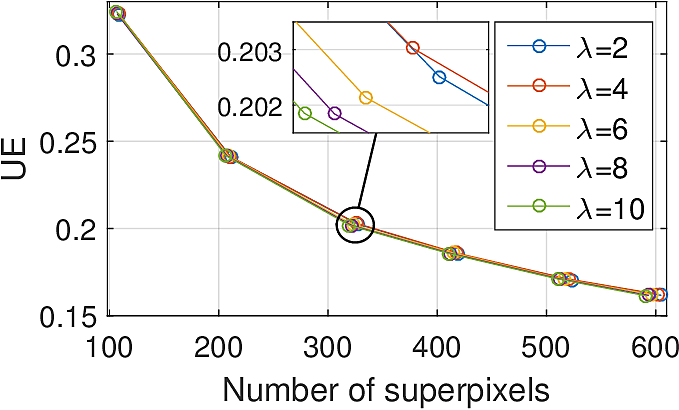}\label{fig:lue}} \quad
\subfloat[]{\includegraphics[width=0.31\linewidth]{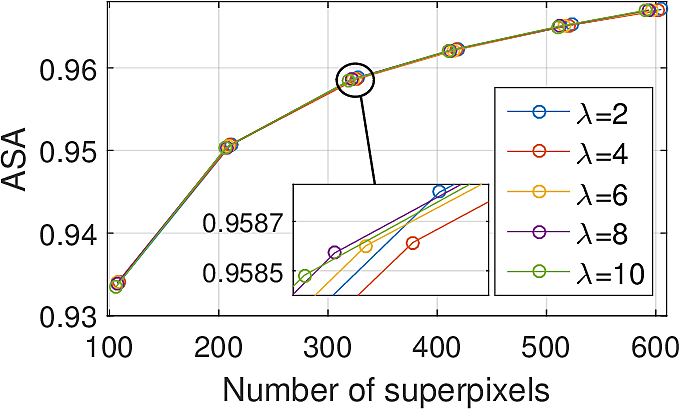}\label{fig:las}}
\caption{Effect of different $\lambda$. 
Experiments are performed on BSDS500 to generate different number of superpixels by adjusting $K$ or $v_x$ and $v_y$, 
and results are averaged over 500 images. 
The results of BR, UE, and ASA are correspondingly plotted in \protect\subref{fig:lbr}, 
\protect\subref{fig:lue}, and \protect\subref{fig:las}. 
In order to see more details, part of the results are zoomed in. (better see in color)}
\label{fig:difflambda}
\end{figure*}

\begin{figure*}[!tb]
\centering
\subfloat[]{\includegraphics[width=0.19\linewidth]{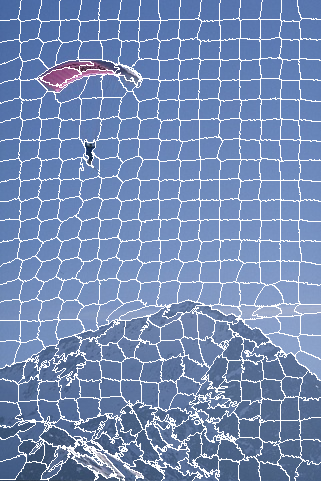}\label{fig:lambda2}}\ 
\subfloat[]{\includegraphics[width=0.19\linewidth]{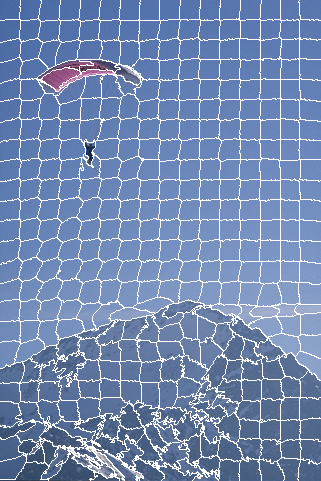}\label{fig:lambda4}}\ 
\subfloat[]{\includegraphics[width=0.19\linewidth]{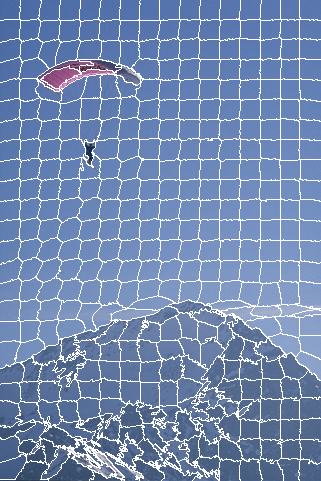}\label{fig:lambda6}}\ 
\subfloat[]{\includegraphics[width=0.19\linewidth]{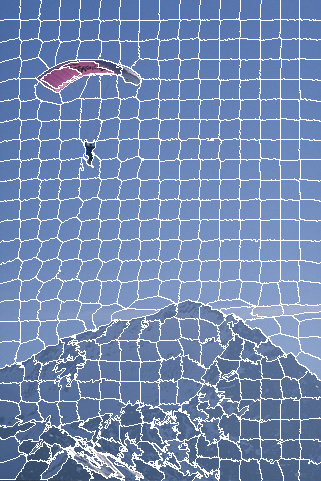}\label{fig:lambda8}}\ 
\subfloat[]{\includegraphics[width=0.19\linewidth]{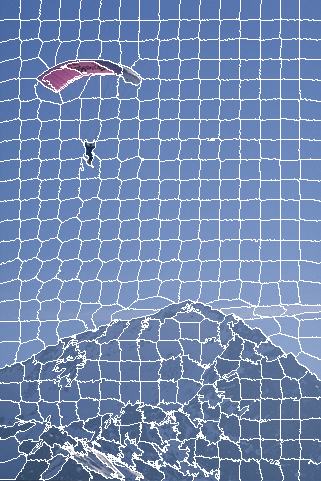}\label{fig:lambda10}}
\caption{visual results with 
\protect\subref{fig:lambda2} $\lambda=2$, 
\protect\subref{fig:lambda4} $\lambda=4$,
\protect\subref{fig:lambda6} $\lambda=6$,
\protect\subref{fig:lambda8} $\lambda=8$, and
\protect\subref{fig:lambda10} $\lambda=10$. 
The test image is from BSDS500 and approximately 400 superpixels are extracted in each image.
}\label{fig:vlambda}
\end{figure*}

\section{Experiment} \label{sec:exp}
In this section, algorithms are evaluated in terms of accuracy, computational efficiency, 
and visual effects. Like many state-of-the-art superpixel algorithms, we also use 
CIELAB color space for our experiments because it is perceptually uniform for small color distance.

\emph{Accuracy}: three commonly used metrics are adopted: 
boundary recall (BR), under-segmentation error (UE), and achievable segmentation accuracy (ASA).
To assess the performance of the selected algorithms, experiments are conducted on the Berkeley Segmentation
Data Set and Benchmarks 500 (BSDS500) which is an extension of BSDS300. These two data sets have been wildly 
used in superpixel algorithms. BSDS500 contains 500 images, and each one of them has the size of 
481$\times$321 or 321$\times$481 with at least four ground-truth human annotations.
\begin{enumerate}

\item[\textbullet] BR measures the percentage of ground-truth boundaries correctly recovered by the superpixel boundary pixels. A true boundary pixel is considered to be correctly recovered if it falls within two pixels from at least one superpixel boundary. A high BR indicates that very few true boundaries are missed.

\item[\textbullet] A superpixel should not cross ground-truth boundary, or, 
in other words, it should not cover more than one object. To quantify this notion, UE calculates the percentage of superpixels that have pixels ``leak'' from their covered object as shown in equation \eqref{eq:ue}.
\begin{equation}\label{eq:ue}
UE = (-1) + \frac{1}{N}\sum_{|s_k \cap s_g|\ >\ \epsilon|s_k|} |s_k|,
\end{equation}
where $s_k$ and $s_g$ are pixel sets of superpixel $k$ and ground-truth segment $g$.
$\epsilon=0.05$ is generally accepted.

\item[\textbullet] If we assign every superpixel with the label of a ground-truth segment into which the most pixels of the superpixel fall, how much segmentation accuracy can we achieve, or how many pixels are correctly segmented? ASA is designed to answer this question. Its formula is defined in equation \eqref{eq:asa} in which $G$ is the set of ground-truth segments.
\begin{equation}\label{eq:asa}
ASA = \frac{1}{N} \sum_{k\in\mathcal{K}} \max\bigg\{|s_k \cap s_g|\ \big|\ g \in G\bigg\}.
\end{equation}
\end{enumerate}

\emph{Computational efficiency}: execution time is used to quantify this property.

\begin{figure*}[!tb]
\centering
\subfloat[]{\includegraphics[width=0.31\linewidth]{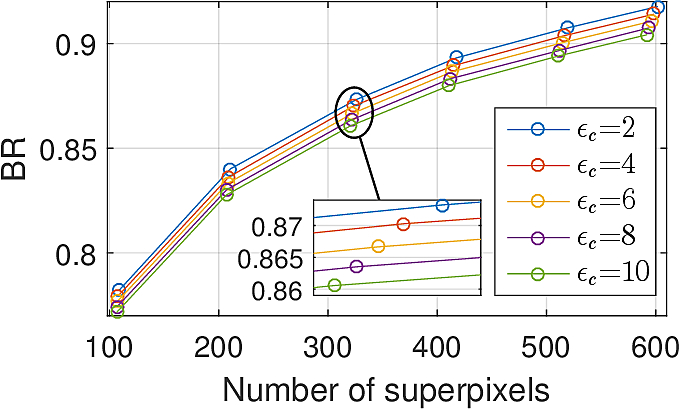}\label{fig:ecbr}} \quad
\subfloat[]{\includegraphics[width=0.31\linewidth]{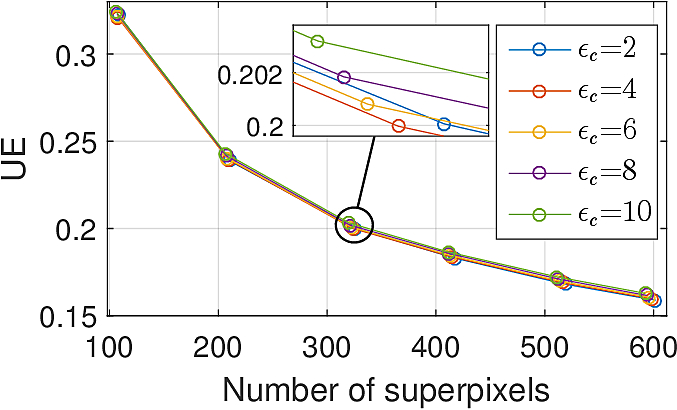}\label{fig:ecue}} \quad
\subfloat[]{\includegraphics[width=0.31\linewidth]{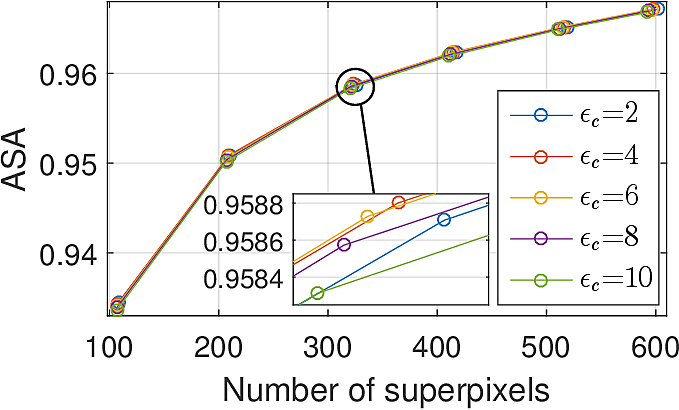}\label{fig:ecas}}
\caption{Results with different $\epsilon_c$. 
Experiments are performed on BSDS500 to generate different number of superpixels by 
adjusting $K$ or $v_x$ and $v_y$, and results are averaged over 500 images. 
The results of BR, UE, and ASA are correspondingly plotted in \protect\subref{fig:ecbr}, 
\protect\subref{fig:ecue}, and \protect\subref{fig:ecas}. 
In order to see more details, part of the results are zoomed in. (better see in colour)}
\label{fig:diffec}
\end{figure*}

\begin{figure*}[!tb]
\centering
\includegraphics[width=0.19\textwidth]{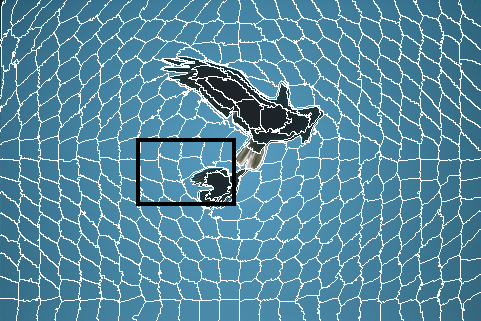}\ %
\includegraphics[width=0.19\textwidth]{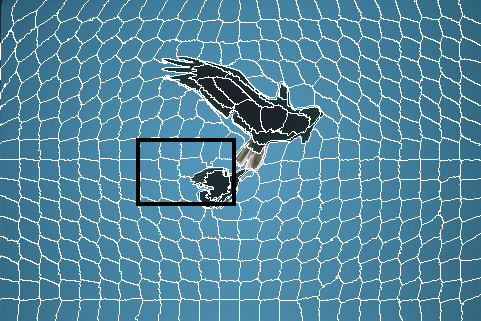}\ %
\includegraphics[width=0.19\textwidth]{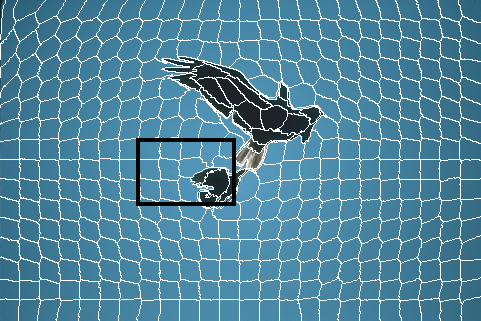}\ %
\includegraphics[width=0.19\textwidth]{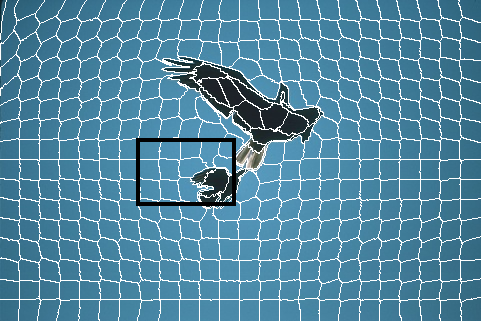}\ %
\includegraphics[width=0.19\textwidth]{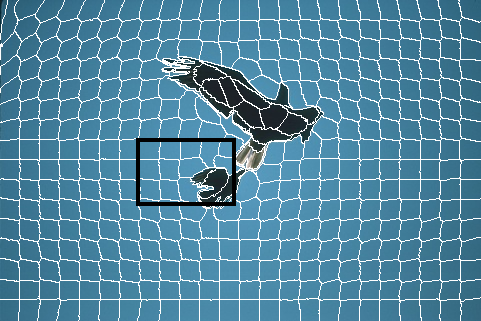}\\%
\vspace{-2.5mm}%
\subfloat[]{\includegraphics[width=0.19\textwidth]{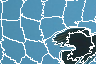}\label{fig:ec2}}\ %
\subfloat[]{\includegraphics[width=0.19\textwidth]{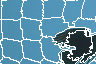}\label{fig:ec4}}\ %
\subfloat[]{\includegraphics[width=0.19\textwidth]{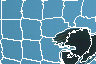}\label{fig:ec6}}\ %
\subfloat[]{\includegraphics[width=0.19\textwidth]{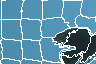}\label{fig:ec8}}\ %
\subfloat[]{\includegraphics[width=0.19\textwidth]{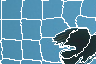}\label{fig:ec10}}%
\caption{visual results with 
\protect\subref{fig:ec2} $\epsilon_c=2$, 
\protect\subref{fig:ec4} $\epsilon_c=4$,
\protect\subref{fig:ec6} $\epsilon_c=6$,
\protect\subref{fig:ec8} $\epsilon_c=8$, and
\protect\subref{fig:ec10} $\epsilon_c=10$. 
The test image is from BSDS500 and approximately 400 superpixels are extracted in each image.
The second row is enlarged from the rectangular marked in the first row.
}
\label{fig:vdiffec}
\end{figure*}

\subsection{Effect of $\lambda$ and $\epsilon_c$}

As shown in Fig. \ref{fig:difflambda}, there is no obvious regularity for the effect of $\lambda$.
In Fig. \ref{fig:difflambda}, the maximum difference between two lines is around 0.001$\sim$0.006 which is very small. 
Although it seems that small $\lambda$ will lead to a better BR result, it is not true for UE and ASA. 
For instance, in the enlarged region of Fig. \ref{fig:lue}, the result of $\lambda=10$ is slightly better than $\lambda=6$. 
Visual results with different $\lambda$ are plotted in Fig. \ref{fig:vlambda}, it is hard for human to distinguish the difference among the five results.

$\epsilon_c$ can be used to control the regularity of the generated superpixels. 
As shown in Fig. \ref{fig:diffec}, small difference of $\epsilon_c$ does not present obvious 
variation for UE and ASA, but it does affect the results of BR. 
In other words, a small variation of $\epsilon_c$ affects the boundary of the produced
superpixels much more than the content of the produced superpixels. 
Generally, a larger $\epsilon_c$ leads to more regular superpixels whose boundary is more smooth. 
Conversely, the shape of superpixels generated with a smaller $\epsilon_c$ 
is relative irregular (see Fig. \ref{fig:vdiffec}). 
Because superpixels with irregular shape will produce more boundary pixels, 
the result of BR with small $\epsilon_c$ is better than that with greater $\epsilon_c$.

We will use $\lambda=8$ and $\epsilon_c=8$ in the following experiments. 
Although this setting does not give the best performance in accuracy, the shape of superpixels using this setting is regular and visually pleasant (see Fig. \ref{fig:vdiffec}\subref{fig:ec8}). 
Moreover, it is enough to outperform state-of-the-art algorithms as shown in Fig. \ref{fig:metrics}.

\subsection{Parallel scalability}
In order to evaluate scalability for the number of processors,
we test our implementation on an machine attached with an Intel(R) Xeon(R) CPU E5-2620 v3 @ 2.40GHz and 8 GB RAM.
The source code is not optimized for any specific architecture. 
Only two OpenMP directives are added for the updating of $\bm\Sigma_k$, 
$\bm{\mu}_k$, and $\bm{R}$, as they can be computed independently (see section \ref{sec:analysis}). 
As listed in Table \ref{tab:para}, for a given image, multiple cores will present a better performance. 

\begin{table}[!hbt]
\caption{run-time (ms) of our implementation on different images with various resolution. The program is executed using 1, 2, 4, and 6 cores.}
\label{tab:para}
\centering
\begin{tabular}{lllll}
\hline\noalign{\smallskip}
Resolution & 1 core  & 2 cores & 4 cores & 6 cores\\
\hline\noalign{\smallskip}
240$\times$320  & 393.646  & 303.821  & 227.078 & 200.708\\
320$\times$480  & 776.586  & 589.785  & 400.073 & 321.548\\
480$\times$640  & 1569.74  & 1011.62  & 743.629 & 624.561\\
640$\times$960  & 3186.71  & 2244.12  & 1353.72 & 1069.79\\
\noalign{\smallskip}\hline
\end{tabular}
\end{table}

\subsection{Comparison with state-of-the-art algorithms}

We compare the proposed algorithm to eight state-of-the-art superpixel segmentation algorithms including 
LSC\footnote{http://jschenthu.weebly.com/projects.html} \cite{LSC},
SLIC\footnote{http://ivrl.epfl.ch/research/superpixels} \cite{SLIC}, 
SEEDS\footnote{http://www.mvdblive.org/seeds/} \cite{seeds15}, 
ERS\footnote{https://github.com/mingyuliutw/ers} \cite{ERS}, 
TurboPixels\footnote{http://www.cs.toronto.edu/~babalex/research.html} \cite{TP}, 
LRW\footnote{https://github.com/shenjianbing/lrw14} \cite{LRW}, 
VCells\footnote{http://www-personal.umich.edu/~jwangumi/software.html} \cite{VCells}, 
and Waterpixels\footnote{http://cmm.ensmp.fr/~machairas/waterpixels.html} \cite{Waterpixels}. 
The results of the eight algorithms are all generated from implementations provided by the authors 
on their respective websites with their default parameters except for the desired number of superpixels, 
which is decided by users.

\begin{figure*}[!ht]
\centering
\subfloat[]{\includegraphics[width=0.31\linewidth]{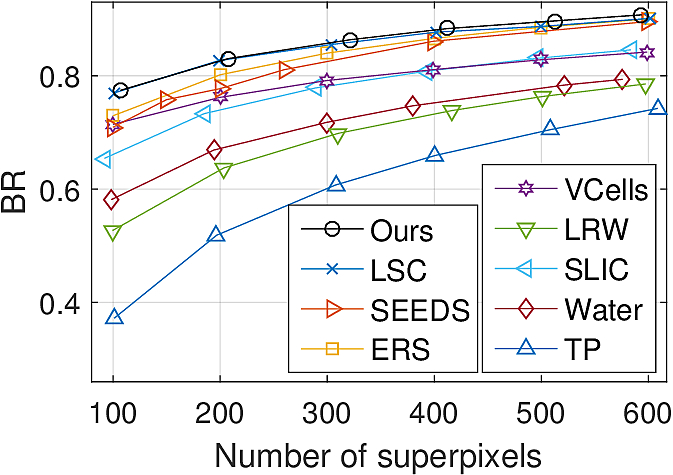}\label{fig:abr}} \quad
\subfloat[]{\includegraphics[width=0.31\linewidth]{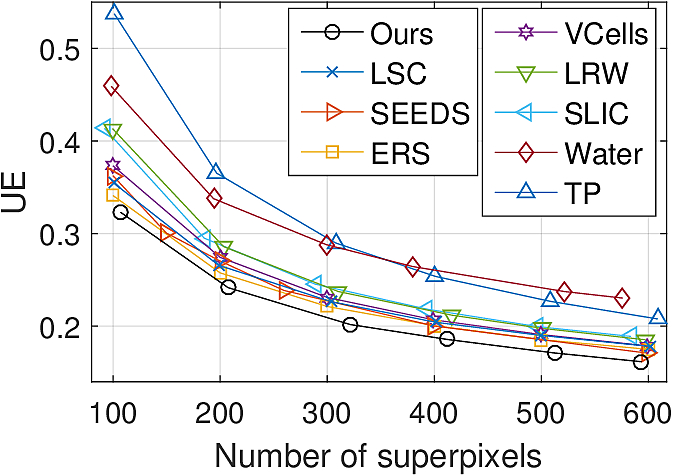}\label{fig:aue}} \quad
\subfloat[]{\includegraphics[width=0.31\linewidth]{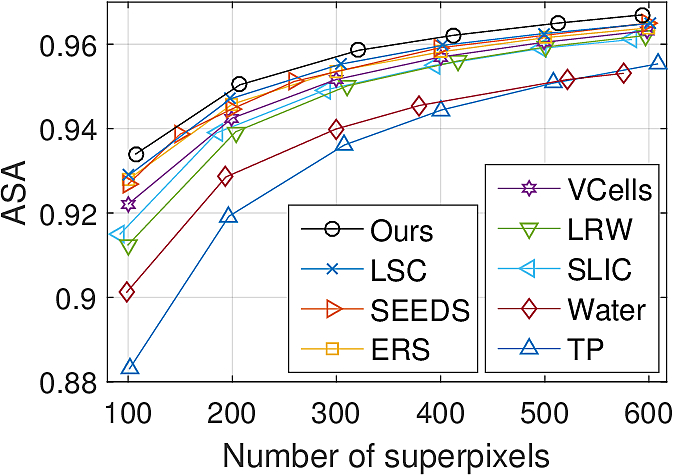}\label{fig:aas}}
\caption{Comparison with state-of-the-art algorithms. 
Experiments are performed on BSDS500  to generate different number of superpixels by adjusting the desired number of superpixels,
and results are averaged over 500 images. 
The results of BR, UE, and ASA are correspondingly plotted in \protect\subref{fig:abr}, \protect\subref{fig:aue}, and \protect\subref{fig:aas}.}
\label{fig:metrics}
\end{figure*}

\begin{figure}
\centering
\includegraphics[width=0.8\columnwidth]{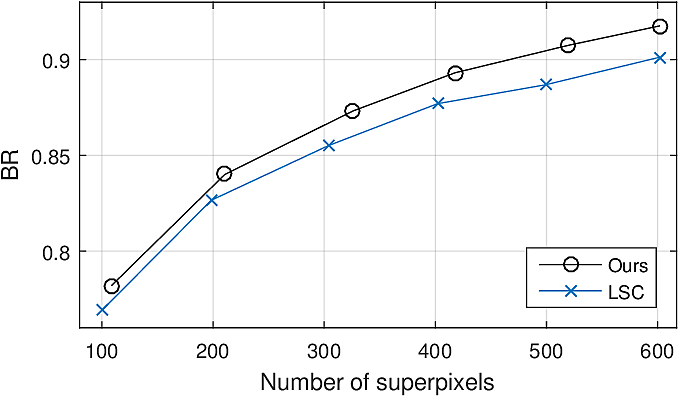}
\caption{Comparison of BR between LSC and our method. Without changing the default value of other parameters in our method, we use $\epsilon=2$ in this figure.}
\label{fig:lscbr}
\end{figure}

As shown in Fig. \ref{fig:metrics}, our method outperforms the selected state-of-the-art algorithms especially for UE and ASA. 
It is not easy to distinguish between our result and LSC in Fig. \ref{fig:metrics}\subref{fig:abr}.
However, if we use $\epsilon_c=2$, our result will obviously outperforms LSC as displayed in Fig. \ref{fig:lscbr}.

\begin{figure}
\centering
\subfloat[]{\includegraphics[width=0.45\linewidth]{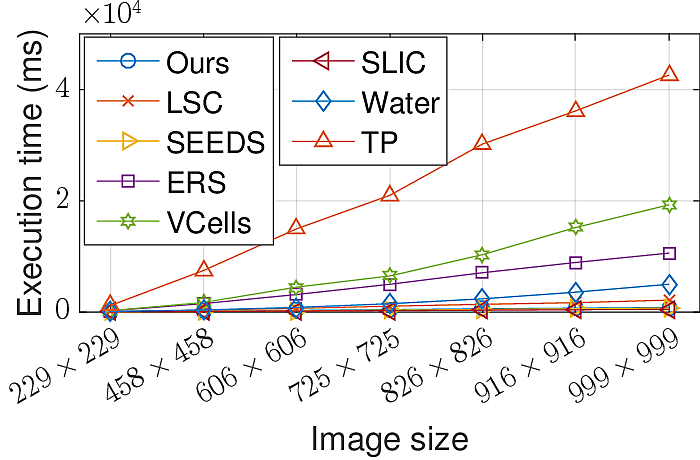}\label{fig:t7}}\
\subfloat[]{\includegraphics[width=0.45\linewidth]{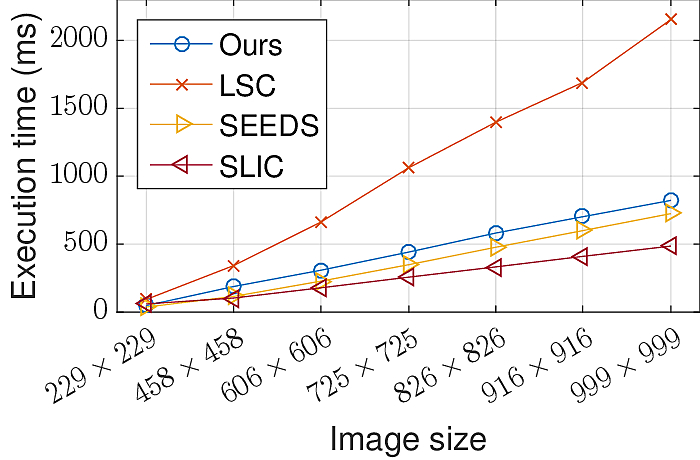}\label{fig:t4}}
\caption{Comparison of run-time. Seven algorithms are compared in \protect\subref{fig:t7}. In order to see more details, the rum-time of the fastest four algorithms is plotted in \protect\subref{fig:t4}.
LRW is not included in the two figures due to its slow speed.
}
\label{fig:time}
\end{figure}

To compare the run-time of the selected algorithms, we test them on a desktop machine equipped with an
Intel(R) Core(TM) i5-4590 CPU @ 3.30GHz and 8 GB RAM. The results are plotted in Fig. \ref{fig:time}. 
According to Fig. \ref{fig:time}\subref{fig:t4}, as the size of the input image increases, run-time of our algorithm grows linearly, which proves our algorithm is of linear complexity experimentally.

\begin{figure*}
\centering
\parbox[b][0.1271\textwidth][c]{2mm}{\rotatebox{90}{\footnotesize{Ours}}}\ %
\includegraphics[height=0.1251\textwidth]{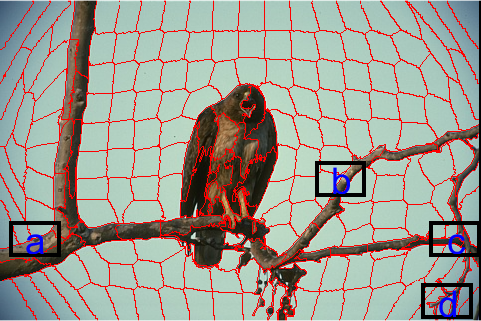}\ %
\includegraphics[height=0.1251\textwidth]{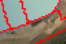}\ %
\includegraphics[height=0.1251\textwidth]{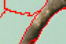}\ %
\includegraphics[height=0.1251\textwidth]{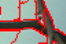}\ %
\includegraphics[height=0.1251\textwidth]{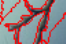}\\%
\vspace{1mm}%
\parbox[b][0.1271\textwidth][c]{2mm}{\rotatebox{90}{\footnotesize{LSC}}}\ %
\includegraphics[height=0.1251\textwidth]{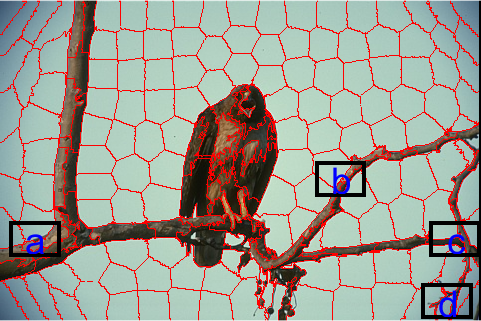}\ %
\includegraphics[height=0.1251\textwidth]{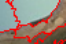}\ %
\includegraphics[height=0.1251\textwidth]{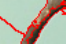}\ %
\includegraphics[height=0.1251\textwidth]{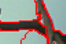}\ %
\includegraphics[height=0.1251\textwidth]{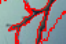}\\%
\vspace{1mm}%
\parbox[b][0.1271\textwidth][c]{2mm}{\rotatebox{90}{\footnotesize{SEEDS}}}\ %
\includegraphics[height=0.1251\textwidth]{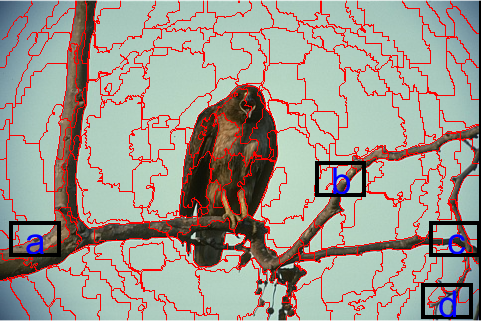}\ %
\includegraphics[height=0.1251\textwidth]{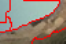}\ %
\includegraphics[height=0.1251\textwidth]{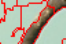}\ %
\includegraphics[height=0.1251\textwidth]{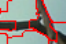}\ %
\includegraphics[height=0.1251\textwidth]{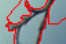}\\%
\vspace{1mm}%
\parbox[b][0.1271\textwidth][c]{2mm}{\rotatebox{90}{\footnotesize{ERS}}}\ %
\includegraphics[height=0.1251\textwidth]{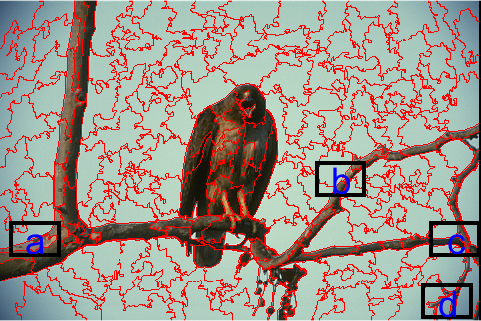}\ %
\includegraphics[height=0.1251\textwidth]{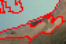}\ %
\includegraphics[height=0.1251\textwidth]{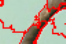}\ %
\includegraphics[height=0.1251\textwidth]{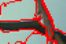}\ %
\includegraphics[height=0.1251\textwidth]{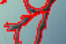}\\%
\vspace{1mm}%
\parbox[b][0.1271\textwidth][c]{2mm}{\rotatebox{90}{\footnotesize{VCells}}}\ %
\includegraphics[height=0.1251\textwidth]{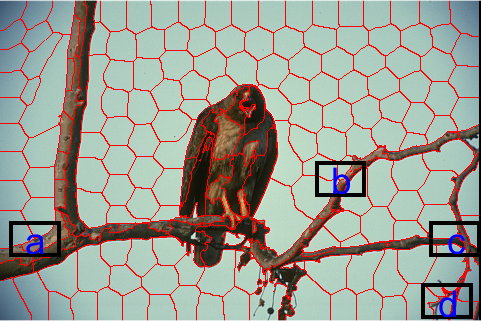}\ %
\includegraphics[height=0.1251\textwidth]{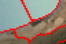}\ %
\includegraphics[height=0.1251\textwidth]{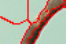}\ %
\includegraphics[height=0.1251\textwidth]{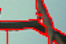}\ %
\includegraphics[height=0.1251\textwidth]{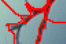}\\%
\vspace{1mm}%
\parbox[b][0.1271\textwidth][c]{2mm}{\rotatebox{90}{\footnotesize{LRW}}}\ %
\includegraphics[height=0.1251\textwidth]{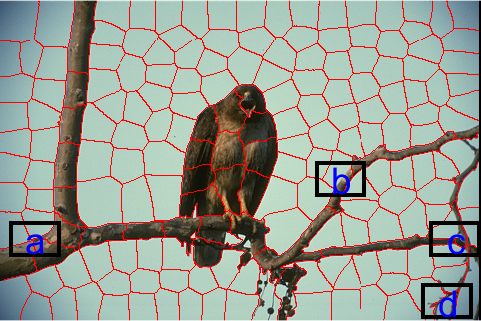}\ %
\includegraphics[height=0.1251\textwidth]{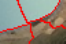}\ %
\includegraphics[height=0.1251\textwidth]{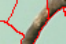}\ %
\includegraphics[height=0.1251\textwidth]{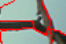}\ %
\includegraphics[height=0.1251\textwidth]{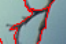}\\%
\vspace{1mm}%
\parbox[b][0.1271\textwidth][c]{2mm}{\rotatebox{90}{\footnotesize{SLIC}}}\ %
\includegraphics[height=0.1251\textwidth]{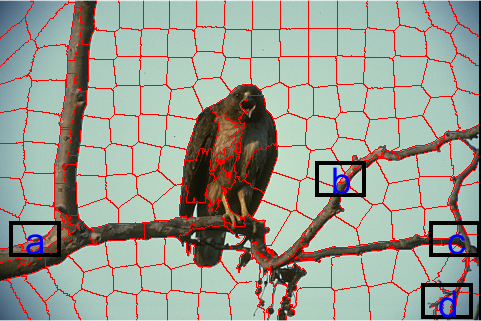}\ %
\includegraphics[height=0.1251\textwidth]{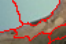}\ %
\includegraphics[height=0.1251\textwidth]{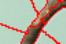}\ %
\includegraphics[height=0.1251\textwidth]{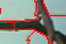}\ %
\includegraphics[height=0.1251\textwidth]{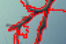}\\%
\vspace{1mm}%
\parbox[b][0.1271\textwidth][c]{2mm}{\rotatebox{90}{\footnotesize{Water}}}\ %
\includegraphics[height=0.1251\textwidth]{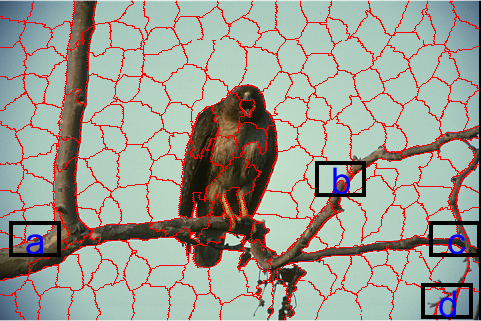}\ %
\includegraphics[height=0.1251\textwidth]{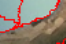}\ %
\includegraphics[height=0.1251\textwidth]{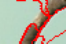}\ %
\includegraphics[height=0.1251\textwidth]{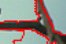}\ %
\includegraphics[height=0.1251\textwidth]{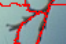}\\%
\vspace{1mm}%
\parbox[b][0.1271\textwidth][c]{2mm}{\rotatebox{90}{\footnotesize{TP}}}\ %
\includegraphics[height=0.1251\textwidth]{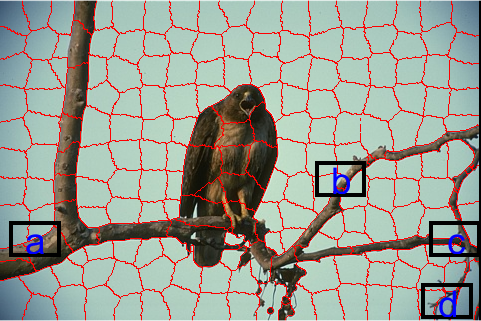}\ %
\subfloat[]{\includegraphics[height=0.1251\textwidth]{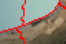}\label{fig:va}}\ %
\subfloat[]{\includegraphics[height=0.1251\textwidth]{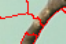}\label{fig:vb}}\ %
\subfloat[]{\includegraphics[height=0.1251\textwidth]{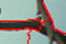}\label{fig:vc}}\ %
\subfloat[]{\includegraphics[height=0.1251\textwidth]{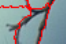}\label{fig:vd}}%
\caption{Visual comparison. The test image is selected from BSDS500. Each algorithm extracts approximately 200 superpuxels. For each segmentation, four parts are enlarged to display more details.}
\label{fig:visual}
\end{figure*}

A visual comparison is displayed in Fig. \ref{fig:visual}. 
According to the zooms, only the proposed algorithm can correctly reveal the segmentations. 
Our superpixel boundaries can adhere object very well.
LSC gives a really competitive result, however there are still parts of the objects being under-segmented.
The superpixels extracted by SEEDS and ERS are very irregular and their sizes vary tremendously.
The remaining five algorithms can generate regular superpixels, but they adhere object boundaries poorly.

\section{Conclusion} \label{sec:cons}

This paper presents an alternative method for superpixel segmentation
by associating each superpixel to a Gaussian distribution with unknown parameters; 
then constructing a Gaussian mixture model for each pixel; 
and finally the superpixel label of a pixel is determined by a posterior probability 
after that the unknown parameters are estimated by the proposed 
algorithm derived from the expectation-maximization method.
The main difference between the traditional GMM method and the proposed one is 
that data points in our model are not assumed to be identically distributed.
Another important contribution is the application of eigendecomposition used in the updating of covariance matrices.

The proposed algorithm is of linear complexity, which has been proved by both theoretical analysis and experimental results. 
What's more, it can be implemented using parallel techniques, and its run-time scales with the number of processors. 
The comparison with the state-of-the-art algorithms shows that the proposed algorithm outperforms the selected methods 
in accuracy and presents a competitive performance in computational efficiency.

As a contribution to open source society, we will make our test code public available at \url{https://github.com/ahban}.




\newpage
\begin{IEEEbiography}[{\includegraphics[width=1in,height=1.25in,clip,keepaspectratio]{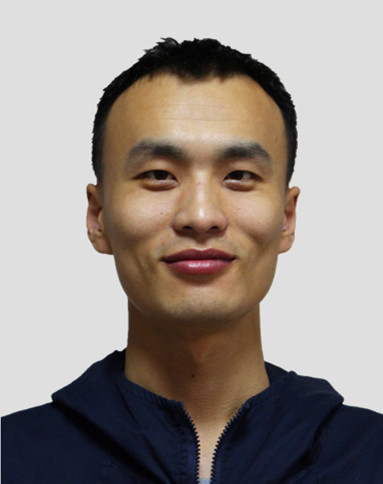}}]
{Zhihua Ban}
received the B.S. degree from China University of Petroleum, Qingdao, China, in 2012. He is currently
pursuing the Ph.D. degree with the State Key Lab for Multispectral Information Processing Technology,
School of Automation, Huazhong University of Science and Technology.
His research interests are in the areas of clustering, image segmentation, and parallel computing.
\end{IEEEbiography}
\begin{IEEEbiography}[{\includegraphics[width=1in,height=1.25in,clip,keepaspectratio]{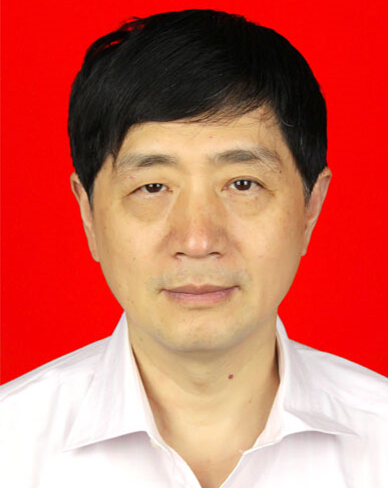}}]
{Jianguo Liu}
received his B. S. degree in mathematics from the Wuhan University of Technology in 1982 and M. S. 
degree in computer science from the Huazhong University of Science and technology in 1984, and Ph. D. 
degree in electrical and electronic engineering from the University of Hong Kong in 1996, respectively. 
He was a visiting scholar with the medical image processing group of the department of radiology at the 
University of Pennsylvania from December, 1998 to July, 2004.  He is currently a professor of School of 
Automation at Huazhong University of Science and Technology in China.  His interests include signal 
processing, image processing, parallel algorithm and structure, and pattern recognition.
\end{IEEEbiography}
\begin{IEEEbiography}[{\includegraphics[width=1in,height=1.25in,clip,keepaspectratio]{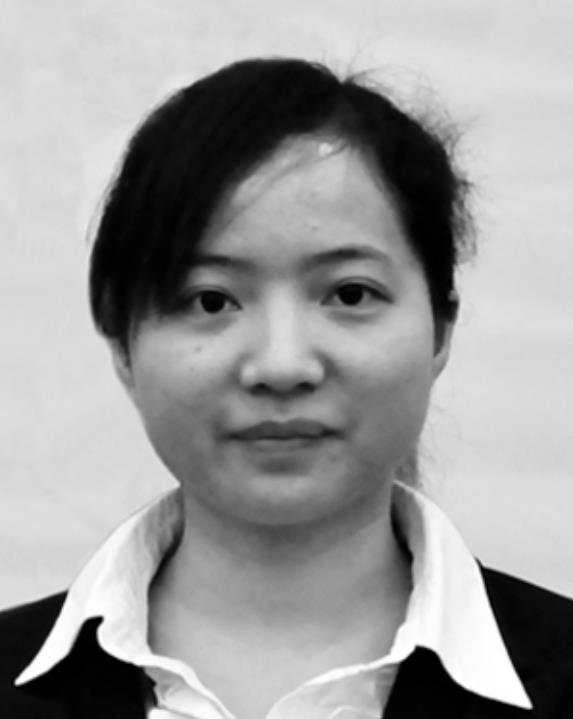}}]
{Li Cao}
received the B. S. degree in Electronic Information Science and Technology from 
the Central South University in 2010, and took successive postgraduate and doctoral programs 
of study in Huazhong University of Science and Technology. She was a visiting PhD student with 
the medical image processing group of the department of radiology at the University of Pennsylvania from 
August, 2014 to September, 2015. She is currently a PhD student of School of Automation at Huazhong 
University of Science and Technology in China. 
Her interests include signal processing, image processing, parallel algorithm and structure.
\end{IEEEbiography}
\vfill

\end{document}